%% file: nfa_icmla.tex
\documentclass[conference]{IEEEtran}
\IEEEoverridecommandlockouts
\usepackage{cite}
\usepackage{amsmath,amssymb,amsfonts}
\usepackage{dsfont}
\usepackage{array}
\usepackage{algorithmic}
\usepackage{graphicx}
\usepackage{textcomp}
\usepackage{xcolor}
\usepackage[caption = false]{subfig}
\usepackage{soul}
\usepackage{bbding}
\usepackage{booktabs}
\usepackage{tabularx}

\def\BibTeX{{\rm B\kern-.05em{\sc i\kern-.025em b}\kern-.08em
    T\kern-.1667em\lower.7ex\hbox{E}\kern-.125emX}}
\begin{document}

\title{A Multi-Scale A Contrario method for Unsupervised Image Anomaly Detection

\thanks{This work was partially funded by a graduate scholarship from Agencia Nacional de Investigación e Innovación, Uruguay.}
}

\author{\IEEEauthorblockN{Matías Tailanian}
\IEEEauthorblockA{\textit{Digital Sense \&} \\
\textit{Universidad de la República}\\
Montevideo, Uruguay \\
mtailanian@digitalsense.ai \vspace{-20pt}}
\and
\IEEEauthorblockN{Pablo Musé}
\IEEEauthorblockA{\textit{IIE, Facultad de Ingenier\'ia} \\
\textit{Universidad de la República}\\
Montevideo, Uruguay \\
pmuse@fing.edu.uy \vspace{-20pt}}
\and
\IEEEauthorblockN{Álvaro Pardo}
\IEEEauthorblockA{\textit{Universidad Cat\'olica del Uruguay} \\
\textit{Digital Sense}\\
Montevideo, Uruguay \\
apardo@ucu.edu.uy \vspace{-20pt}}
}

\maketitle

\begin{abstract}
Anomalies can be defined as any non-random structure which deviates from normality. Anomaly detection methods reported in the literature are numerous and diverse, as  what is considered anomalous usually varies depending on particular scenarios and applications. In this work we propose an \textit{a contrario} framework to detect anomalies in images applying statistical analysis to feature maps obtained via convolutions. We evaluate filters learned from the image under analysis via patch PCA, Gabor filters and the feature maps obtained from a pre-trained deep neural network (Resnet). The proposed method is multi-scale and fully unsupervised and is able to detect anomalies in a wide variety of scenarios. While the end goal of this work is the detection of subtle defects in leather samples for the automotive industry, we show that the same algorithm achieves state of the art results in public anomalies datasets. 
\end{abstract}


\section{Introduction}

Anomaly detection is an active field that has been studied for decades, motivated by a wide variety of practical applications ranging from robotics and security to health care. 
One of the most relevant applications is automatic or aided product quality inspection in industrial production. 

In most situations it is very difficult or even impossible to collect a statistically relevant sample of all types of anomalies, given its rare nature and intra-class variability. In addition, in most cases anomalies are very subtle, and the line that separates them from normality is thin and fuzzy. These kinds of problems are typically formulated as one-class classification problems~\cite{oneclass}, where the strategy is based on learning and characterizing the statistics of normality. The key advantage of this scheme is that only ``normal'' (i.e. anomaly free) samples are needed for the training process. Then, new tested samples are assigned an anomaly score that represents the deviation of the tested sample to the characterized normality.

\input{plot_example_images}

A very different situation arises when normality and abnormality are not absolute notions, but relative to the sample itself. Indeed, in many scenarios, a very same pattern can be considered normal or abnormal when it occurs within two different samples. This situation is depicted in Figure~\ref{fig:example-images}, which shows some examples of the data considered for the application that motivates the present work. In this application, which will be described in detail in Section~\ref{sec:bader}, the goal is to segment defects on processed leather pieces for the automotive industry. The leather could be coloured with a wide variety of tones, may present different textures, and the strength of the texture engraving may also differ. Note for example that the anomalous pattern in Figure~\ref{fig:example-images}(b) should be considered normal if it was present in the sample of Figure~\ref{fig:example-images}(a). 

In summary, the problem we aim to tackle is such tat normality can only be modeled from a single image sample, and where we have no prior knowledge whether this sample contains anomalies or not. These conditions frame our solution to a small set of methods, discarding all techniques based on learning from any set of images, and those that extract statistics from a set of anomaly-free samples.


A well established methodology to deal with unsupervised anomaly detection under these conditions is the \textit{a contrario} approach~\cite{lowe2012perceptual,desolneux2007gestalt}. This methodology is commonly used in anomaly detection and has proven to produce impressive results in many tasks, such as clustering, edges and line segments detection~\cite{cao2008theory, von2008lsd}
, general point alignments~\cite{lezama2014contrario}, among others. 
In this work we propose to apply this methodology to detect anomalies, after a step of feature extraction devised to enhance the difference between normal and anomalous patches in the image (Section~\ref{sec:method}).

\input{table_related_work}

Based on the fact that learning a statistical model of anomalies 
is not possible due to their very limited number of occurrences, the {\em a contrario} methodology, as other approaches, focuses on the design of a {\em background model} or null hypothesis that characterizes normality. Anomalies are then detected as events such that the probability of occurrence under the background model is so small that they are much more likely to result from another cause. It is also interesting to note that in most scenarios, the number of observed anomalies is so limited that the background model for normality can be learnt from the whole image (there is no need to separate samples into normal and abnormal). What makes the {\em a contrario} framework different and particularly useful, is that it automatically fixes detection thresholds that control the number of false alarms (NFA)~\cite{desolneux2007gestalt}
, allowing not only to detect rare events in very diverse backgrounds but also to associate a rareness score. This score has a clear statistical meaning: it is an estimate of the number of occurrences of an observed event if it was produced by the background model. Therefore, if the background model is accurate, setting this threshold to 1 ensures that all detections are not realizations of the background process.

In short, in this work we develop a fully unsupervised multi-scale \textit{a contrario} anomaly detection method that proves to be highly versatile with very good results. Our algorithm does not need any normal or anomalous datasets, nor any priors. In practice, this method is parameterless and benefits from all the computational power of neural networks libraries.

The remainder of this paper is organized as follows: in Section~\ref{sec:relatedwork} we 
present an exhaustive review of state of the art methods and compare their key elements. Then, we present our work as a general algorithm to detect anomalies, explaining the method and showing results in Sections~\ref{sec:method},~\ref{sec:filtering}~and~\ref{sec:results}. 
Finally in Section~\ref{sec:bader} we apply the proposed method to our particular industrial application, and we present its results.

\section{Related Work}
\label{sec:relatedwork}

There is a wide literature on anomaly detection. 
The most common approach to detect anomalies consists in modeling the distribution of normal samples by means of a background model of randomness. The design of this model is usually problem-specific. For instance, 
in~\cite{text} the authors characterize normality by fitting different GMMs with dense covariance matrix to a 4-level image pyramid. As previously mentioned, the {\em a contrario} approach is based on the same principle, but differs from the others in the sense that it allows to derive detection thresholds by estimating and controlling the expected number of occurrences of an event under the background model. 
An excellent example is the work by Ehret et al.~\cite{ehret2019reduce}, in which they eliminate the self-similarity of the image by averaging the most frequent patches using Non Local Means
, and perform a detection using the NFA score over the residual image, which is formed by noise and anomalies. In this way, the general background modeling problem gets reduced to a noise modeling problem, making the algorithm to work with any kind of background.

More recently, 
new data-driven methods based on deep neural networks (DNN) were developed. In most cases they use only anomaly-free images, achieving impressive results in publicly available anomaly datasets. A comparison between some key elements of different state of the art works is presented in Table~\ref{tab:related-work}. From this table is easy to see that most of the methods operate with some kind of supervision, which is the main difference of our proposed algorithm. A common classification is to separate methods in two categories: \textit{feature-similarity based} and \textit{reconstruction based} methods.

\textbf{Feature-similarity based methods} Works based on feature similarity start building a new feature space, where it is claimed to be easier to identify anomalies. The second step is usually to define and measure some notion of similarity, which can range from a simple kNN or some probability distribution fitting, to some learned metric, ad hoc to the specific problem. 
For example in~\cite{spade}, the authors construct a pyramid of features using the activation maps of a Wide-ResNet50, pre-trained on ImageNet, and use these feature maps for finding the K nearest anomaly-free images. 
In~\cite{padim} excellent results are reported by modeling normality with a multivariate Gaussian distribution for each patch position, using the output of a pre-trained CNN, and measuring the Mahalanobis distance to normality for each patch. 
In \cite{fcdd} a fully convolutional neural network is trained, where the output features preserve the spatial information, and are indeed a down-sampled version of an anomaly heatmap. By using an HSC (Hyper-Sphere Classifier) loss, the nominal samples are encouraged to be mapped near the center of the new space, and the anomalous samples away. The network is trained with both normal and anomalous samples. Anomalous samples could be synthetic, but authors found that using even a few examples of real labeled anomalies the method perform much better. In~\cite{dict} the authors extract the features from a pre-trained DNN and build a dictionary over the features, which is subsequently used to measure the distance of the target image to normality. A student-teacher framework is used in~\cite{student-teacher}, where the student learns the distribution of anomaly-free images by matching their features with the teacher network accordingly. The anomaly score is obtained by comparing the output of both networks. In \cite{patch-svdd} the idea of \textit{Deep-SVDD}~\cite{deep-svdd} is extended by constructing a hierarchical encoding of the image patches. 
The authors show that the performance of the whole algorithm is improved by adding a self-supervised learning term to the loss function. 
The hard boundary of SVDD associated with the hyper-sphere can cause the model to overfit the training data. Therefore, in~\cite{gsvdd} the authors propose to replace it by a Gaussian SVDD (GSVDD), and treat its mean and co-variance as latent variables to be estimated. Driven by the adversarially constrained auto-encoder interpolation, 
they make the distribution of the normal samples denser, and thus reduce the likelihood of an anomalous image embedding laying between normal samples.

\textbf{Reconstruction-based} Generative models, as VAEs~\cite{kingma2013auto} or GANs~\cite{goodfellow2014generative} are based on the idea that the network trained only on non-defective samples, will not be able to reconstruct the anomalies at test time. In practice, they usually present high capacity and generalization power, leading to even a good reconstruction of the anomalies. In~\cite{intra} the authors address this issue by stating the generative part as an inpainting problem. 
As stated in~\cite{bergmann-ssim}, better results could be obtained measuring the reconstruction error with the structural similarity measure (SSIM). 
In~\cite{cutpaste} the authors design a special classification proxy task in order to be able to learn deep representation in a self-supervised manner, by training the network with random cut and pasted regions over anomaly-free images. 
In~\cite{dfr} the authors propose to train an Auto-Encoder over the feature maps of a pre-trained network. Instead of reconstructing the image itself, the network performs a deep features reconstruction. 
Unlike AEs and VAEs, which learn a direct representation from the image to the latent code, GANs learn the mapping from the latent space to realistic normal images. As the latent space transitions are smooth, in~\cite{anogan} the authors propose an iterative method to find a latent code that generates a similar image to the target, enabling to find anomalies by looking at the reconstruction error. 

When learning from one-class data, a usual approach consists in mapping normal images near some centroid in the latent space, and expect an anomalous image to lay far from this centroid. To overcome this potential issue, a common strategy is to design some proxy task and train a network using a self-supervised approach. For example, in~\cite{rotnet} the authors present a distribution-augmented contrastive learning technique via data augmentation, that obstructs the uniformity of contrastive representations, making it easier to isolate outliers from inliers.

\begin{figure*}[t]
    \subfloat{\includegraphics[width=1\linewidth]{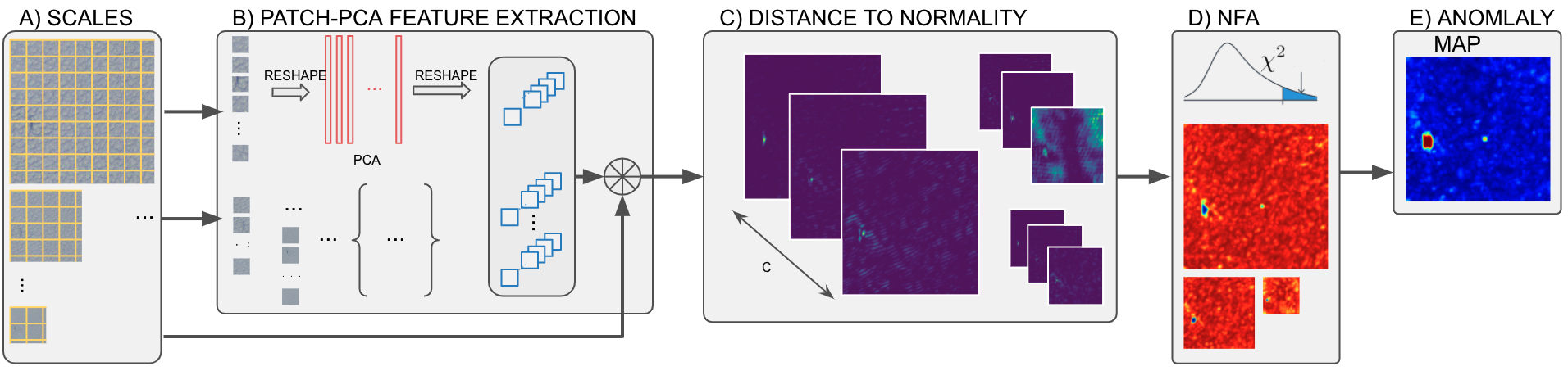}} \hfill
\\[-20px]
    \caption{Method diagram. Our method can be divided in 5 stages. Stage A) consists in creating a pyramid of images of different scales. In stage B) we obtain the filters by performing a Patch-PCA transformation, and use them to filter the input image. At the beginning of stage C) we have one feature map for each filter and for each input channel, and we combine them by computing one mahalanobis distance for each channel. Stage D) performs the computation of the Number of False Alarms. The Stages B), C) and D) are computed for each scale independently, and all results are merged in stage E).}
    \vspace{-10pt}
    \label{fig:method-diagram}
\end{figure*}

As we are specially interested in methods that can be used with no training and over a wide variety of different images as input, we also focus on the domain of \textit{saliency detection}. For this work we are only interested in textured images, where the images are uniform. In these cases, from the perception point of view, any structure or pattern that deviates from normality, should also be the most salient part of the image. There is an extensive literature dedicated to it. To cite a few relevant works, a united approach to the activation and normalization steps of saliency computation is introduced in~\cite{gbvs-harel2007graph}, by using dissimilarity and saliency to define edge weights on graphs that are interpreted as Markov chains. In~\cite{drfi-jiang2013salient} the salient object detection is formulated as a regression problem in which the task is learning a regressor that directly maps the regional feature vector to a saliency score. 
In~\cite{salicon-huang2015salicon}, the authors propose to estimate saliency by combining multiple popular DNN architectures for object recognition 
that are known to encode powerful semantic features. 

Although there is a vast literature related to anomaly detection, we found that only few algorithms are able to perform the detection with no prior knowledge of the specific task/problem, and most of them try to model normality from a set of normal images, often needing them to be aligned and acquired in very controlled situations. In this work we develop a simple and effective unsupervised algorithm that only uses the input image, with no need of any prior knowledge or training data, and estimates a ``rareness score'' given by the Number of False Alarms (NFA) for each pixel/region of the image. 

\section{Method}
\label{sec:method}

The kind of anomalies could be very diverse. In this work, we concentrate in low-level anomalies (by opposite of semantic anomalies). 
Using classic image processing techniques we achieved a simple, fast, and yet very effective method that has proven to work successfully in very different kinds of images, as will be shown in Section~\ref{sec:results}. As this method was developed to solve an industrial problem, both accuracy and speed are of major importance.

The 
proposed method consists of five stages: A) Image pyramid generation by scale decomposition; B) Patch-PCA Feature Extraction; C) Mahalanobis distance to normality over the activation maps; D) NFA computation; and E) Generation of the final anomaly map, by merging the results of different scales. The method is summarized in Figure~\ref{fig:method-diagram}, and each of the stages are explained in the following. 

\subsection{Image decomposition: scale's pyramid}
Since anomalies can occur at any scale, it is crucial to detect them no matter their size. This is performed by means of a multi-scale approach based on an image pyramid where each new scale is half the width and the height of the previous one. In what follows we explain the procedure for one scale.

\subsection{Patch-PCA Feature Extraction}

To represent the local statistics of the pixels in the acquired images, such as its surrounding texture, we associate to each pixel a patch of $s \times s$ pixels centered on it. We treat each image channel separately, so for each channel $c$, each pixel location is characterized by a $n$-feature vector, where $n = s \times s$. 
We apply PCA to decorrelate the patch feature vector and treat the new coordinates independently.   

PCA performs a linear transformation, $ \mathbf{y}_i = P^T \mathbf{x}_i$, where $\mathbf{x}_i \in \mathds{R}^n$ and $i \in \{1\dots N\}$ is a set of observations (patches), $ \mathbf{y}_i \in \mathds{R}^m$ with $m\leq n$ their projections, and the columns $\{\mathbf{p}_i\}_{i=1}^n$ of $P$ the PCA eigenvectors. The element $j$ in vector $\mathbf{y}_i$ is the result of the inner product between $\mathbf{x}_i$ and $\mathbf{p}_j$. Since these projections are inner products over patches, they can be efficiently computed as 2D convolutions with the eigenvectors as kernels, using common deep learning libraries. Extracting the PCA eigenvectors can also be seen as a methodology for finding the filters used to extract features, based only on the input image. This characteristic is of crucial importance in our application, as it defines specific adapted filters for each case. Once we have obtained the filters, we convolve the input image by channel with each filter (e.g. project into the PCA space), obtaining a descriptor of size $m$ for each of the $c$ channels, for each pixel.

\subsection{Distance to normality}
\label{sec:mahalanobis}

From now on we consider the first $m < n$ PCA components, i.e., a number of PCA projections smaller than dimension of the original space. 
Taking into account that vectors $\mathbf{p}_i$ are orthogonal, the squared Euclidean distance between two reconstructed patches $\mathbf{\hat x^a}$ and $\mathbf{\hat x^b}$ can be computed from their projections and the mean $\mathbf{ \bar{x} }$ as:
$$ d^2(\mathbf{\hat x^a}, \mathbf{\hat x^b}) = d^2 \big( \mathbf{\bar x} + \sum_i a_i \mathbf{p}_i, \mathbf{\bar x} + \sum_i b_i \mathbf{p}_i \big) = \sum_i(a_i-b_i)^2. $$
To compute the distance to normality, we need a set of normal samples and its corresponding projections. Assuming the size of the anomaly is small compared to the size of the image and does not affect the statistics, we can obtain a set of normal samples from the image itself. We denote this set as $\{\mathbf{x^b}\}$, and its projections to the PCA space as $\{\mathbf{b}\}$. Estimating the mean $\mathbf{\mu}$ and variance $\Sigma$ parameters of the distribution of normal samples, and recalling that the PCA components are uncorrelated ($\Sigma = \mbox{diag}(\sigma_1^2,...,\sigma_n^2)$ is a diagonal matrix) we obtain that the squared Mahalanobis distance of a given patch to normality is given by
\begin{equation*}
   d^2(\mathbf{\hat x^a}, \textit{normality}) = \sum_i \frac{(a_i - \mu_i)^2}{\sigma_{i}^2}.
   \label{ec:mahalanobis}
\end{equation*}

In the PCA formal approach, we must build the normal set by subtracting the mean of all patches. Instead, we use an approximation of the mean, considering just the mean of the whole image. As all patches are extracted from the same $c$-channel images, this difference is small, and it does not affect the Mahalanobis distance as the terms get cancelled. 

Note that the considered PCA components are obtained by filtering each of the $c$ channels with $m$ filters. For each channel, the response of all filters are added, leading to a total of $c$ images of Mahalanobis distances.

Empirically it was observed that these images of PCA projections have a Gaussian distribution (we will come back to this point in the following section).
Assuming independence (the projections are already uncorrelated), this implies that each of the $c$ images of the squared Mahalanobis distances follows a $\chi^2$ distribution with $m$ degrees of freedom. 
%

\subsection{Number of False Alarms}
\label{sec:nfa}

The \textit{a contrario} model is based on the rejection of a null-hypothesis. We consider our null-hyphotesis ($\mathcal{H}_0$) to be normality, namely, the absence of any anomaly. The idea behind the NFA computation is to detect any configuration of pixels that are unlikely to happen under $\mathcal{H}_0$, thus probably caused by another cause, and therefore anomalous. The NFA value itself has a very intuitive interpretation: if a certain tested pattern has a value of $\text{NFA}=1$ (log-NFA=0), it means that this pattern can occur at most 1 time under $\mathcal{H}_0$ (i.e. caused by our model of normality).
The NFA represents an upper-bound on the expectation in an image~\cite{desolneux2007gestalt} (here the Mahalanobis distance image) of the number of pixels/regions of probability less than the one of the considered sample. Formally, a function $\phi$ is a NFA if 
$\forall \delta > 0, \quad \mathds{E} [\# \{ \phi(Z_k) \leq \delta \} ] \leq \delta$, 
where $Z_k,\; k \in \{1,\dots, N_T\}$ is a set of random variables that satisfy the null-hypothesis. The function $\phi$ guarantees a bound of the expectation of the number of false alarms. Namely, by thresholding $\phi(Z_k)$ by $\delta$, we should obtain up to $\delta$ false alarms when $Z_k$ verifies the null-hypothesis. In our case $\{Z_k\} = \left \{ D_{i,j}, \; (i,j) \in \{1,\dots,W\} \times \{1,\dots,H\} \right \}$, where $D_{i,j}$ is the value of the squared Mahalanobis distance image at pixel $(i,j)$. The hypothesis of normality $\mathcal{H}_0$ is that $D_{i,j} \sim \chi ^2 (m) $ with $m$ the number of filters, and our function $\phi$ is the probability $N_T \; \mathbb{P}(D_{i,j} > d_{i,j})$ computed over the Mahalanobis distance to normality, where $d_{i,j}$ is the actual measured value, and $N_T$ the number of executed tests. Instead of  assuming  that the  projections onto  the  PCA  space follow  a  Gaussian  distribution, we  could  use  an  empirical  distribution,  but  the  theoretical distribution makes the threshold to be robust and constant for all images, avoiding computation time and bringing an elegant and  stable  theoretical  bound.

We present two different versions of the NFA computation: by pixels and by blocks. Both strategies generate one NFA map per channel and per scale. It is usually the case where the anomaly only shows in one particular channel (e.g. some color anomaly). Therefore, in order to keep all possible detections, we keep the minimum of all NFAs across all scales. The output of this stage is one NFA map for each scale.\\


\vspace{-3pt}
\noindent
{\bf NFA by pixels} As explained in Section~\ref{sec:mahalanobis}, if we assume the distribution of the PCA-projected images to be Gaussian, we obtain a Mahalanobis distance following a $\chi^2$ distribution with as many degrees of freedom as the number of components we used in the PCA ($m$). We empirically observed this fact both in the coefficients of the PCA transformation, and on the Mahalanobis distance used for the NFA analysis. 

Since we test all pixels of the image we set $N_T=HW$, with $H$ and $W$ the image dimensions. Therefore, in the case of pixel evaluation, the NFA function $\phi$ is simply 
\vspace{-3pt}
$$\text{NFA} = \phi(D_{i,j}) = HW \, \mathbb{P}_{\chi ^2(m)}(D_{i,j}>d_{i,j}).$$
\noindent
{\bf NFA by blocks} This algorithm begins with the extraction of candidate regions for each channel, by thresholding each squared Mahalanobis distances 
$D_{i,j}$ with a threshold $\tau$, which can be easily set. Since 
the $D_{i,j}$ are assumed to follow a $\chi^2(m)$ distribution, we set $\tau$ such as the corresponding $p$-value is $p=0.01$.
Under the null-hypothesis $\mathcal{H}_0$, all these pixels are uniformly distributed on the whole image. Following the idea in~\cite{marina}, we base this NFA criterion on the detection of suspicious concentration of these candidates. Considering a block $B$ of size $w \times w$, and the set of candidates  
$\mathcal{L}_B=\{(i,j) \in B \text{ s.t. } D_{i,j} > \tau\}$ within the block, 
we define a new random variable 
$U_{i,j}$
that takes value $1$ if 
$(i,j) \in \mathcal{L}_B$ and $0$ otherwise. 

To search for unusual concentrations of candidate pixels, we need to evaluate the probability $P(U \geq |\mathcal{L}_B |)$, with $U=\sum_{i,j} U_{i,j}$. Evaluating this probability is not straightforward, as the $U_{i,j}$ are not independent. To overcome this issue, we consider $s^2$ separate tests, one for each possible sub-sampled grid by a factor $s$ in both dimensions. For each one we observe $w^2/s^2$ distances. The probability of observing at least $|\mathcal{L}_B|/s^2$ distances greater than $\tau$, among the previous ones, is therefore given by the tail of the binomial law, $\mathcal{B} \left( \frac{w^2}{s^2}, \frac{|\mathcal{L}_B|}{s^2}, p \right)$. 

The number of blocks $B$ to be tested is $HW/w^2$. Testing all $s \times s$ sub-sampled grids per block leads to a total number of $N_T = HW s^2/w^2$ tests. Therefore, the NFA associated to a block $B$ is given by
$$ \text{NFA} = \frac{HW}{w^2} s^2 \mathcal{B} \left( \frac{w^2}{s^2}, \frac{|\mathcal{L}_B|}{s^2}, p \right).$$

\subsection{Final anomaly map}
At this point, we have obtained one NFA map for each scale. The next step is to combine them in order to generate a unique map gathering the information of all scales. To do so, we follow the intuition that if some structure results to be anomalous at a certain scale, it has to be marked as anomalous in the final map. Therefore we compute the final NFA map as the minimum value of NFA of each scale for each pixel. These maps are combined by up-sampling smaller scales to match the original size. Finally, we translate this NFA map into an rareness measure by simply defining the Anomaly Score as $AS=-\log_{10} (\text{NFA})$, which provides a more intuitive map where the larger the value associated to a structure, the higher its  rareness.

\input{table_results.tex}
\section{Alternative filtering}
\label{sec:filtering}
In this section we analyze the effect of using other filters instead of the principal components. We propose two modifications to our method, that substitute the stages A) and B) presented above.

\noindent
{\bf Gabor filters.} Instead of performing the scale decomposition, and computing the Patch-PCA descriptors per image, we propose to use Gabor filters of different sizes. We show in the following section that the results obtained with this modification still achieves a good performance. Furthermore, avoiding the PCA computation significantly reduces the computational cost, which is of major importance for the industrial application we need to solve.\\

\vspace{-5pt}
\noindent
{\bf ResNet feature maps.} Inspired by recent works in anomaly detection, we can substitute the first two stages of the method by a ResNet-50 pre-trained on ImageNet.
We directly obtain the activation maps by feeding this ResNet with the input image, and looking at specific intermediate outputs of the network. Specifically, we keep the output of the last convolution of three layers:  \textit{layer1/conv3, layer2/conv3, layer3/conv3}. As we need the feature maps to be decorrelated for our hypothesis to stand, we then apply a PCA over the feature maps, capturing up to 90\% of the variance. The rest of the method remains unchanged, and as we show in Section~\ref{sec:results}, we obtain a very good performance, and some times even better results.

\begin{figure}
    \centering
    \subfloat{\includegraphics[width=.85\columnwidth]{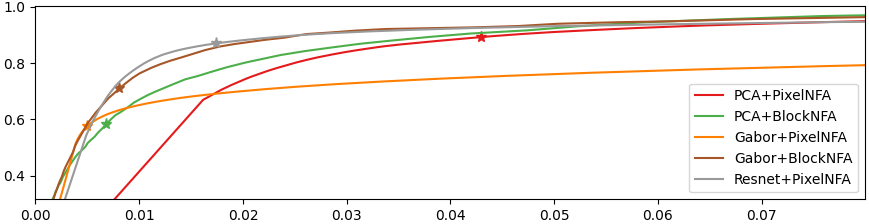}} \hfill
    \\[-5px]
    \caption{ROC for leather subset of MVTec AD, for all variants of the proposed method. The value with AS=0 (NFA=1) is indicated with a star. Note that the point AS=0 lay close to the optimal point of the ROC.}
    \vspace{-8pt}
    \label{fig:roc}
\end{figure}

\begin{figure}
    \centering
    \subfloat{\includegraphics[width=.85\columnwidth]{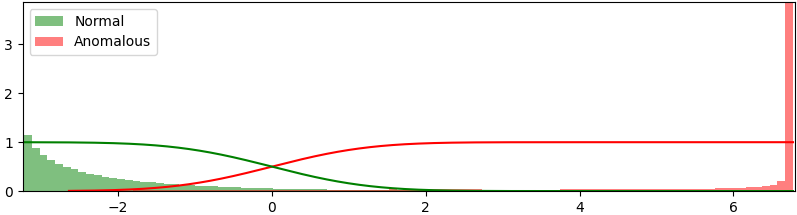}} \hfill
    \\[-8px]
    \caption{Distribution of the Anomaly Score $\text{AS}=-\log(\text{NFA})$ values for normal and anomalous pixels of the leather subset of MVTec AD. overlapped with the cumulative density function for both classes. Note the large gap between classes, and the adequate choice of NFA values close to AS=0 to derive the threshold.}
    \vspace{-10pt}
    \label{fig:gap}
\end{figure}

\section{Results}
\label{sec:results}


To the best of our knowledge, there are few truly unsupervised methods that can work only with the image being inspected, with no prior training or information about normal samples, as shown in Table~\ref{tab:related-work}. In order to compare our proposed method with the state of the art, we use MVTec AD~\cite{mvtec}, a recent anomaly dataset that simulates faults or anomalies in industrial conditions. The MVTec AD dataset contains several subsets, divided in two categories: texture and objects. We focus on the texture category, as we are interested in the detection of low level anomalies. Also, one of the subsets has special importance for us, 
as it considers anomalies in leather samples. MVTec AD also provides several non-faulty images, which most of the state of the art methods use for learning the normality. In order to extend the comparison, we also include the baseline methods, even if they use normal samples and the comparison is not completely fair, in their favor. Results are shown in Table~\ref{tabresultsmvtec}. To perform the comparison, we use the area under the receiver operating characteristic curve (ROC AUC), as it is the most used in the literature, and enable us to compare with state of the art results.

The results in Table~\ref{tabresultsmvtec} corresponding to the Block-NFA version are indicated with a cross; the others use Pixel-NFA. The proposed algorithm achieves state of the art results, comparing it even with some methods that have a clear benefit by using anomaly-free images for training. In some subsets our methods manage to obtain the best results. As mentioned before, we pay special attention to the leather case since it is related to our industrial application (see next section). For this dataset, our approach achieves the highest score.
Furthermore, if we consider the average score over all subsets, all variants of the proposed algorithm perform the best, among which stands out the ResNet variant, that in addition to produce the best results, it is also the fastest variant of our approach. 

As mentioned before, one of the benefits of using the \textit{a contrario} framework is the simplicity to set the threshold for segmentation. As the metric we used to compare with all methods is independent of the threshold, we present the Figure~\ref{fig:roc}, where we show the curves obtained for the leather dataset with all the variants of our proposed algorithm, and indicate with a star the point where AS=0, i.e. NFA=1. These curves does not exactly correspond to the values reported in Table~\ref{tabresultsmvtec}, as we used only two scales, in order to obtain better results specifically for the leather subset. It can be seen that values close to $\text{AS}=0$ lay near the optimal point in the ROC. Also, another interesting verification is to estimate the GAP between the anomaly scores of the faulty and non-faulty regions. In order to do that, we show in Figure~\ref{fig:gap} the distribution of the anomaly scores for all leather samples, overlapped with the cumulative density functions for both classes, making it clear that there is a good separation between them. Additionally we evaluate this GAP analytically by computing the difference between the median values of the anomaly scores $\text{AS}$ of the two classes for all variants, obtaining the following results: \textit{PCA+PixelNFA}: 2.25, 
\textit{PCA+BlockNFA}: 4.06, 
\textit{Gabor+PixelNFA}: 9.12, 
\textit{Gabor+BlockNFA}: 4.62, 
\textit{ResNet+PixelNFA}: 3.15.
In all cases we obtained a very good GAP, and classes well separated. We can observe that depending on the feature extraction used, in one case the AS gap is greater using BlockNFA, and in the other using PixelNFA. For the PCA-based variant we used $m=45$, $s=17$, and 4 scales. For the Gabor variant $m=72$, $s$ ranging from 7 to 31, and 4 scales. The Block NFA is computed considering blocks of size $w=51$ with a stride of 10 pixels.

\begin{figure}
\centering
    \includegraphics[width=.85\columnwidth]{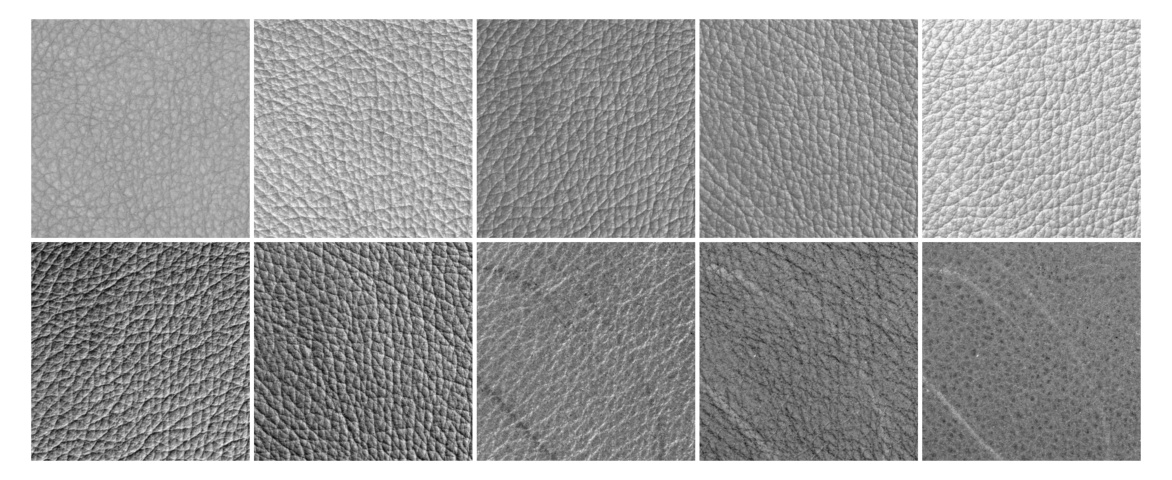}
    \vspace{-10pt}
    \caption{Top row: original images. Bottom row: Output of the preprocessing step. Note how the PCA-based preprocessing better manages to single out potential anomalies.}
    \vspace{-10pt}
    \label{fig:pca_lights_merge}
\end{figure}

\section{Anomaly detection in leather for the automotive upholstery industry}
\label{sec:bader}

In this section we present the industrial problem that motivates this work, carried out for a world leading producer of leather upholstery for the high-end automotive industry.   

The raw skin of animals can present multiple kinds of defects, caused by the living animal itself (e.g. tick, lice and mite marks, scars), 
or in the subsequent processes that transform raw leather into textured leather pieces ready to be installed in a vehicle. Various types of defects (holes, machinery drag, etc.) may be introduced in each step of the process, such as differences in the processes of tanning, differences on the roller pressing for texturing (Figure~\ref{fig:example-images}(a)-(b)), etc. In industries that work with leather, these defects must be detected as soon as possible in the production line to take the corresponding actions (avoiding cutting on the defect, leaving it in inconspicuous areas, or even discarding the pieces). Having a human operator to perform the quality control has several downsides, such as the inability to maintain the necessary attention level for long periods, criterion differences between different operators, and of course the time and workforce costs.

Each vehicle manufacturer allows a small margin of defects in the leather pieces that they buy. When this margin is exceeded the entire batch of leather pieces is returned, incurring in large losses for the producing companies. On the other hand, production is limited by the number of personnel available for inspection. Since defects are searched manually in all samples at various points in the process, this is a bottleneck in the production capacity of the plant.

As the defects are often very hard to see to the naked eye, a dedicated acquisition setup was designed. 
For each skin sample, a total number of 5 images are acquired with a single camera pointing at the sample at nadir, using different lightning conditions and orientations. More precisely, a diffuse light next to the camera and 4 directional grazing lights, located in each corner of the table. A typical set of images acquired with this setup are shown in Figure~\ref{fig:pca_lights_merge}, top row.

\subsection{Preprocessing} 

The main algorithm described above, can be fed either with all 5 raw images or with the images resulting from this pre-processing step. 
Contrarily to the most common use of PCA, where only the first components are kept, in this case we are more interested in the last ones. Indeed, the first components correspond to the directions of largest variance and therefore encode global information of the image, such as color or texture. Since anomalies are particular local structures, they are mostly present in the last components. Figure \ref{fig:pca_lights_merge} depicts a visual example. The first row corresponds to the 5 raw images, and the second row to the PCA components (the first image corresponds to the first eigenvector). This example shows that the anomaly is very hard to visualize in all the raw images, but stands out more clearly in the last 3 components of the projected space. 

\input{plot_bader}

\subsection{Results}
For this industrial application, no reliable anomaly segmentation is available. We only were able to gather a limited number of defective samples, that were annotated coarsely and with notable errors by a operator, during the normal inspection work. We present the ROC AUC results for all our method variants, and some examples for visual inspection in Figure~\ref{fig:results-bader}, showing the input image, its NFA maps and their corresponding segmentation with $NFA=0$ for some variants of the proposed method: \textit{PCA+PixelNFA}, \textit{Gabor+BlockNFA}, and \textit{ResNet+PixelNFA}. With all the variants of our method, we succeed in detecting all defects for this samples. In case of the Block-NFA we obtain a coarser map compared to Pixel-NFA, caused by the computation by blocks. On one hand, this characteristic makes the map cleaner and softer, but in the other hand it may cause the evaluation metric used to drop, as it depends on a fine segmentation of the anomaly. The ROC AUC results obtained are the following: PCA+PixelNFA: 83.03\%, PCA+BlockNFA: 84.20\%, Gabor+PixelNFA: 81.81\%, Gabor+BlockNFA: 85.78\%, ResNet+PixelNFA: 77.00\%. We do not consider the combination ResNet+BlockNFA, as the block is too big compared to the activation maps of the network.

Since we have an input of 5 images (1 diffuse light and 4 grazing directional lights), we cannot directly compare other state of the art methods for this dataset, as they expect a single image as input. Moreover, for the ResNet variant of our method we had to include a modification, feeding the ResNet independently for each image and concatenating the output feature maps. Also, for these examples we use only the output of the first layer of ResNet, \textit{layer1/conv3}.

We tested our algorithm in production with thousands of leather samples in the modality of assessment for the operator. More important than the results commented above, the first results obtained in production demonstrate that this tool provides useful assistance to the the operators, indicating problematic zones and making their work easier and faster.

\section*{Conclusions}

In this work we presented a fully unsupervised {\em a contrario} method for anomaly detection, which aims at detecting defects in leather samples from the automotive industry, achieving industry standards. Although the method was designed for this particular application, its foundations are general enough and proved to perform successfully in a wide variety of defects in images of different scenarios. The proposed approach outperforms other state-of-the-art methods on MVTec dataset, and has proven to work specially well for leather samples. 

Future potential improvements include substituting PCA by non linear transformations obtained from auto-encoders and/or normalizing flows trained on normal data. In this setting, the multi-scale approach may be included inherently in the network by means of a U-shaped network. Another line of work to be explored is the combination of the {\em a contrario} approach with DNNs in order to output detections with a confidence score, or to train the ResNet we utilized with leather anomaly-free samples, in order to obtain a more specific feature extraction for our task.


\bibliographystyle{IEEEtran}
\bibliography{nfa_icmla}

\end{document}

%% file: plot_example_images.tex
\captionsetup{position=top}
\begin{figure}[tb]

\subfloat[]{\includegraphics[width=0.192\linewidth]{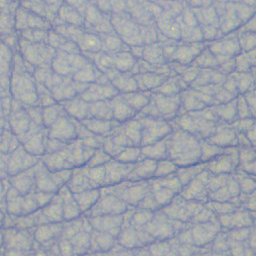}} \hfill
\subfloat[]{\includegraphics[width=0.192\linewidth]{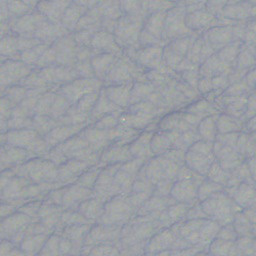}} \hfill
\subfloat[]{\includegraphics[width=0.192\linewidth]{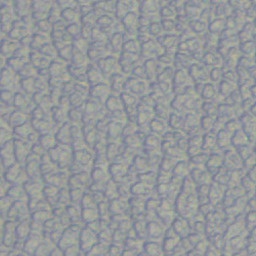}} \hfill
\subfloat[]{\includegraphics[width=0.192\linewidth]{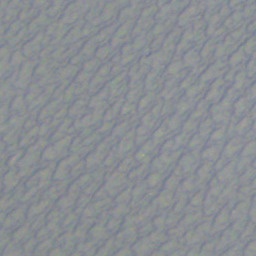}} \hfill
\subfloat[]{\includegraphics[width=0.192\linewidth]{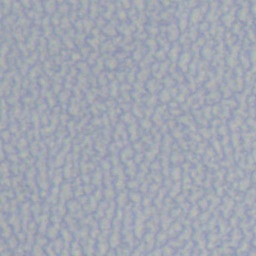}} \hfill \\[-17px]

\captionsetup{position=bottom}
\subfloat[]{\includegraphics[width=0.192\linewidth]{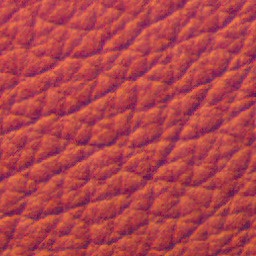}} \hfill
\subfloat[]{\includegraphics[width=0.192\linewidth]{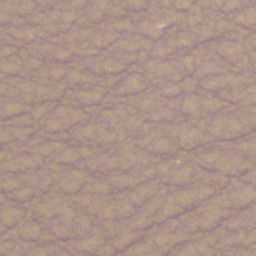}} \hfill
\subfloat[]{\includegraphics[width=0.192\linewidth]{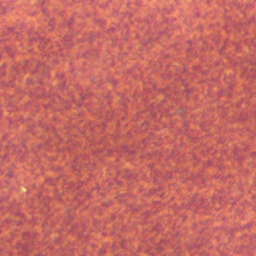}} \hfill
\subfloat[]{\includegraphics[width=0.192\linewidth]{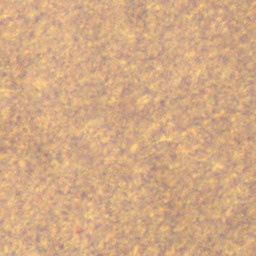}} \hfill
\subfloat[]{\includegraphics[width=0.192\linewidth]{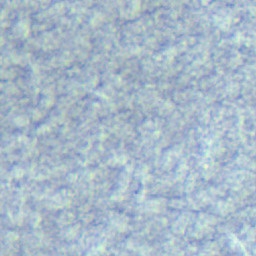}} \hfill \\[-10px]
    \caption{[Best viewed in color] Motivating application: defect detection in leather samples. Example images showing the diversity of samples. Image (a) and (b) correspond to the same texture, with different engraving strength. The defect present in (b) may not be considered an anomaly if it was in (a). Images (c), (d) and (e) are examples of different textures. (f) and (g) show different color possibilities, and (h), (i) and (j) correspond to the down side of leather samples.}
    \vspace{-18pt}
    \label{fig:example-images}
\end{figure}

%% file: table_related_work.tex
\begin{table*}[t]
\caption{Related work comparison showing some of the key elements of each method: how they generate the final anomaly map, whether if they use some pre-trained network or not, how is the usage of normal and abnormal images, the kind of supervision, if they solve a proxy task and the way to achieve the multi-scale analysis.}
\label{tab:related-work}
\vspace{-10pt}
\scriptsize
\begin{center}
    \begin{tabular}{m{7em}|m{17em}|m{3em}|m{5em}|m{3.3em}|m{2em}|m{4.5em}|m{5.5em}|m{9.4em}}
        
        {\bfseries Method} & {\bfseries Anomaly map} & {\bfseries Pre-trained} & {\bfseries Normal images usage} & {\bfseries Anomaly img use}  & {\bfseries Thr} & {\bfseries Supervision} & {\bfseries Proxy task} & {\bfseries Multi-Scale}\\
        \hline  
        
\textbf{TextureInsp}~\cite{text} & Negative log-likelihood for trained GMM & No & Fit GMM & No & No & Sup. & No & 4-layer pyramid \\ 
\textbf{SPADE}~\cite{spade} & Euclidean distance to normal samples & Resnet & kNN & No & No & Sup. & No & DNN scales \\
\textbf{DictSimilarity}~\cite{dict} & Distance to normality & Yes & Build dict. & No & No & Sup. & No & Different patch sizes \\ 
\textbf{FCDD}~\cite{fcdd} & Activation layer from network & No & FCN & Yes & No & Sup. & No & DNN scales \\ 
\textbf{PaDiM}~\cite{padim} & Mahalanobis distance by patch & Resnet & Mah. dist. & No & No & Self-Sup. & No & DNN scales \\
\textbf{PatchSVDD}~\cite{patch-svdd} & Distance to normal samples & No & Encoder & No & No & Self-Sup. & Patch-gather & Arch design \\ 
\textbf{RotDet}~\cite{rotnet} & KDE at each patch location & No & DNN & No & No & Self-Sup. & Rotation & DNN scales \\ 
\textbf{STPM}~\cite{student-teacher} & Student-teacher L2 distance & Resnet & Student net & No & No & Self-Sup. & No & DNN scales \\ 
\textbf{DetInNoise}~\cite{ehret2019reduce} & Log-NFA map & Optional & No & No & Yes & UnSup. & No & Downsampled images \\ 
\hline
\textbf{SSIM-AE}~\cite{bergmann-ssim} & Reconstruction error & No & AE & No & No & Self-Sup. & No & AE architecture \\ 
\textbf{AnoGan}~\cite{anogan} & Reconstruction error & No & GAN & No & No & Self-Sup. & No & GAN architecture \\ 
\textbf{InTra}~\cite{intra} & Reconstruction error & No & Transformer & No & No & Self-Sup. & Inpainting & Downsampled images \\ 
\textbf{CutPaste}~\cite{cutpaste} & One class over representation & No & DNN & No & No & Self-Sup. & Cut-Paste & Replicating \\ 
\textbf{DFR}~\cite{dfr} & Reconstruction error & VGG19 & VAE & No & No & Self-Sup. & No & DNN scales \\ 
\hline 
\textbf{GBVS}~\cite{gbvs-harel2007graph} & Equilibrium distribution of markov chain & No & No & No & No & UnSup. & No & Downsampled images \\ 
\textbf{SALICON}~\cite{salicon-huang2015salicon} & Activation layer from network & No & No & Yes & No & Sup. & No & Arch design \\ 
\textbf{DRFI}~\cite{drfi-jiang2013salient} & RF Regression & No & No & Yes & No & Sup. & No & Multi-Level segmentat. \\ 
\hline 
\textbf{Ours} & Log-NFA map & Optional & No & No & Yes & UnSup. & No & Downsampled images \\ 

    \end{tabular}
\end{center}
\normalsize
\vspace{-15pt}
\end{table*}

%% file: table_results.tex
{\setlength{\tabcolsep}{0.35em} {\renewcommand{\arraystretch}{1.2}%
\begin{table}[t]
\caption{Area under the receiver operating characteristic curve (ROC AUC). In red the best score, blue the second best and green the third. The ranking of the algorithms is done by sorting the performance and summing the positions for each algorithm. The one with lowest value is the best.}\label{tabresultsmvtec}
\vspace{-5pt}
\centering
    \begin{tabular}{l|c|c|c|c|c|c|c}
        {\bfseries Method} & {\bfseries Carpet} & {\bfseries Grid} & {\bfseries Leather} & {\bfseries Tile} & {\bfseries Wood} & {\bfseries Mean} & {\bfseries Rank} \\ 
        \hline
AE-SSIM~\cite{bergmann-ssim} & \textcolor{green}{\textbf{0.87}} & \textcolor{red}{\textbf{0.94}} & 0.78 & 0.59 & 0.73 & 0.78 & 7\\ 
AE-L2~\cite{bergmann-ssim} & 0.59 & 0.90 & 0.75 & 0.51 & 0.73 & 0.70 & 9\\ 
AnoGan~\cite{anogan} & 0.54 & 0.58 & 0.64 & 0.50 & 0.62 & 0.58 & 14\\ 
CNN Dict~\cite{dict} & 0.72 & 0.59 & 0.87 & \textcolor{red}{\textbf{0.93}} & \textcolor{red}{\textbf{0.91}} & 0.80 & 5\\ 
Text insp.~\cite{text} & \textcolor{blue}{\textbf{0.88}} & 0.72 & \textcolor{green}{\textbf{0.97}} & 0.41 & 0.78 & 0.75 & 8\\ \hline
DRFI~\cite{drfi-jiang2013salient} & 0.72 & 0.58 & 0.68 & 0.72 & 0.67 & 0.67 & 10\\ 
GBVS~\cite{gbvs-harel2007graph} & 0.45 & 0.73 & 0.86 & 0.49 & 0.79 & 0.66 & 11\\ 
Salicon~\cite{salicon-huang2015salicon} & 0.53 & 0.79 & 0.76 & 0.40 & 0.72 & 0.64 & 13\\ 
DetNoise~\cite{ehret2019reduce} & 0.57 & 0.68 & 0.81 & 0.53 & 0.57 & 0.63 & 12 \\ \hline
Ours-PCA & 0.76 & 0.88 & \textcolor{blue}{\textbf{0.98}} & 0.69 & \textcolor{green}{\textbf{0.86}} & \textcolor{green}{\textbf{0.83}} & 4\\ 
Ours-PCA~\textsuperscript{\Cross} & 0.65 & \textcolor{green}{\textbf{0.91}} & 0.87 & \textcolor{blue}{\textbf{0.84}} & \textcolor{blue}{\textbf{0.87}} & \textcolor{green}{\textbf{0.83}} & \textcolor{green}{\textbf{3}}\\ 
Ours-Gabor & 0.77 & 0.90 & 0.95 & 0.79 & \textcolor{blue}{\textbf{0.87}} & \textcolor{blue}{\textbf{0.86}} & \textcolor{blue}{\textbf{2}}\\ 
Ours-Gabor~\textsuperscript{\Cross} & 0.69 & 0.83 & 0.92 & \textcolor{green}{\textbf{0.80}} & 0.84 & 0.82 & 6\\ 
Ours-ResNet & \textcolor{red}{\textbf{0.94}} & \textcolor{blue}{\textbf{0.92}} & \textcolor{red}{\textbf{0.99}} & 0.77 & 0.86 & \textcolor{red}{\textbf{0.90}} & \textcolor{red}{\textbf{1}}\\
        \end{tabular}
\vspace{-10pt}
\end{table}
}}

%% file: plot_bader.tex
\captionsetup{position=top}
\begin{figure}[t]

\subfloat{\includegraphics[width=0.195\linewidth]{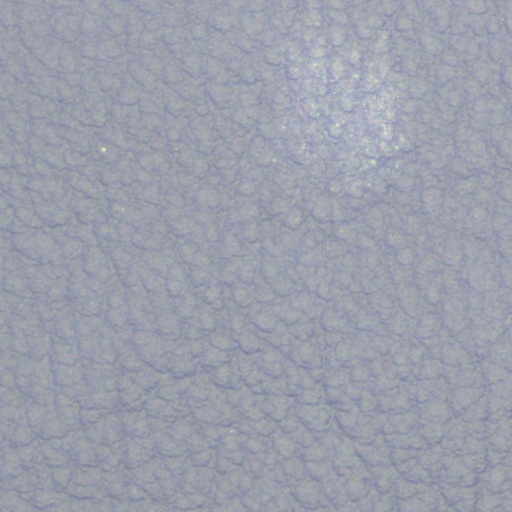}} \hfill
\subfloat{\includegraphics[width=0.195\linewidth]{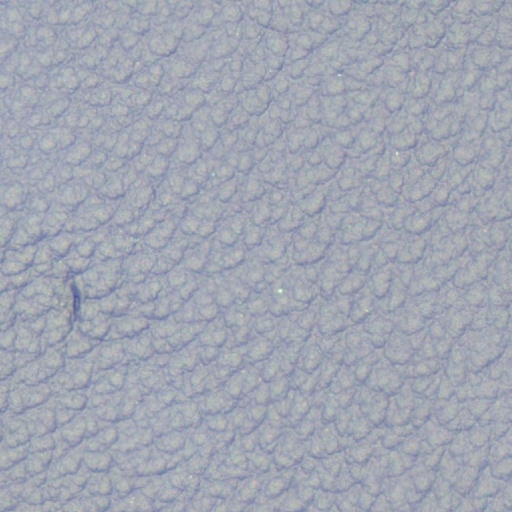}} \hfill
\subfloat{\includegraphics[width=0.195\linewidth]{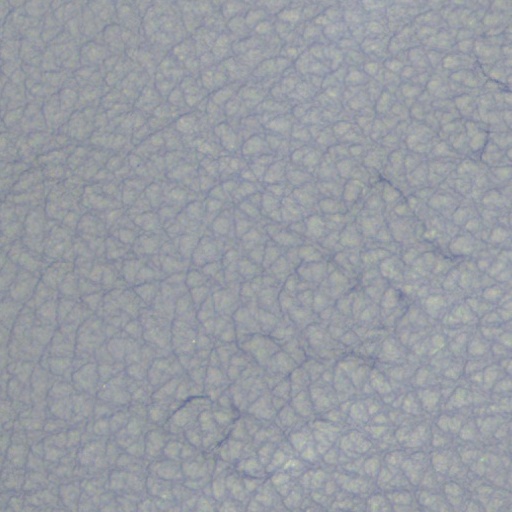}} \hfill
\subfloat{\includegraphics[width=0.195\linewidth]{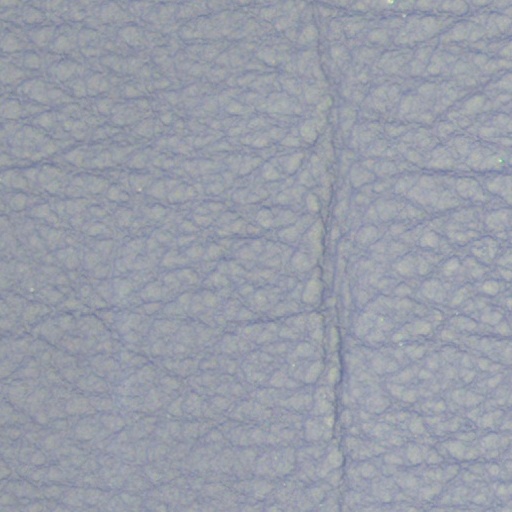}} \hfill
\subfloat{\includegraphics[width=0.195\linewidth]{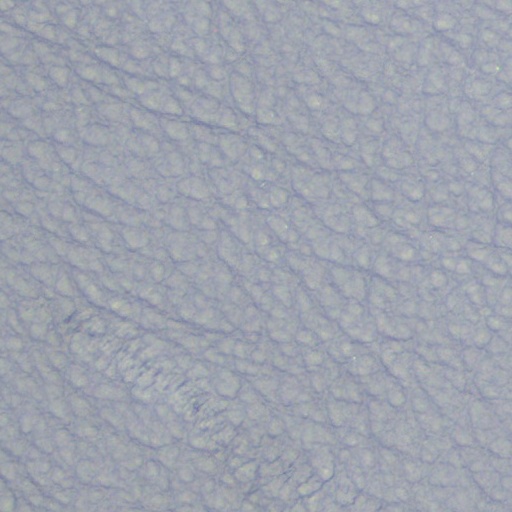}} \hfill \\[-20px]

\subfloat{\includegraphics[width=0.195\linewidth]{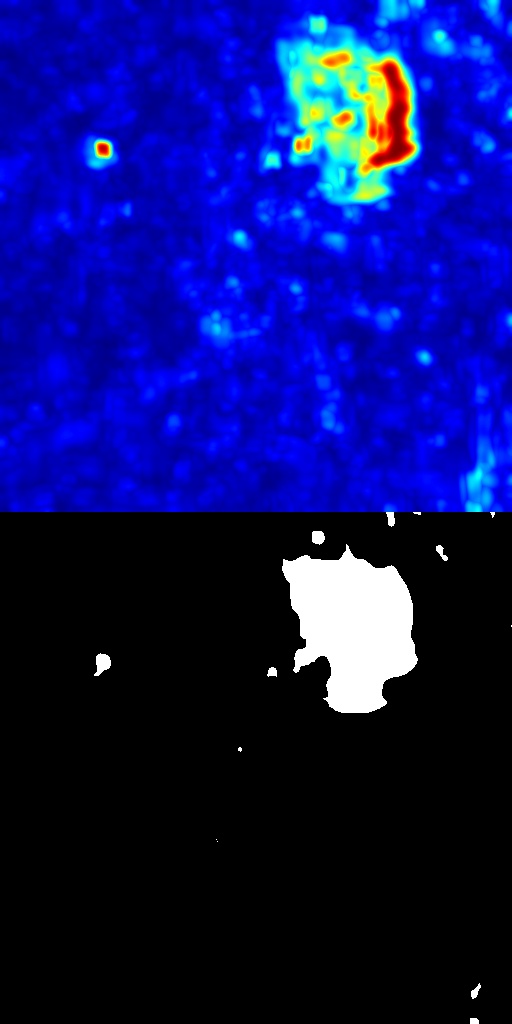}} \hfill
\subfloat{\includegraphics[width=0.195\linewidth]{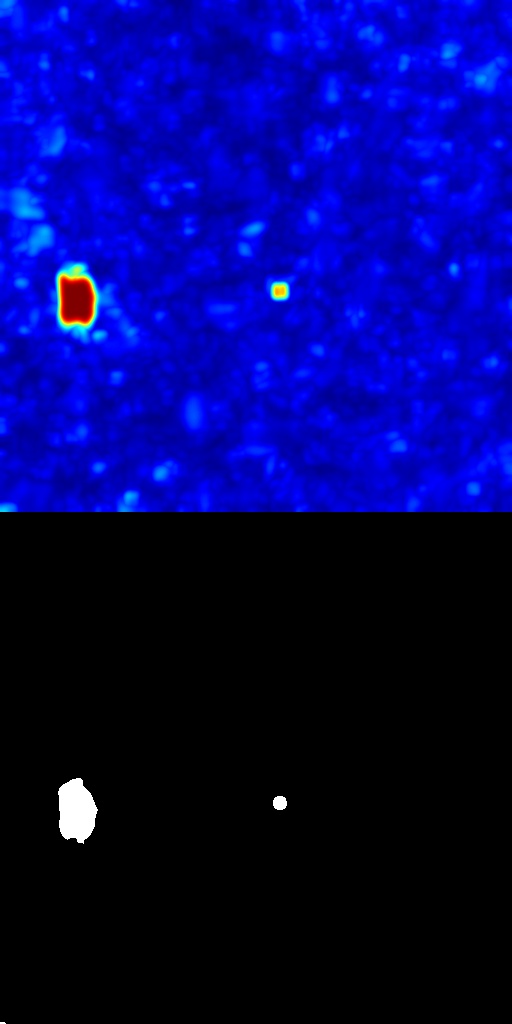}} \hfill
\subfloat{\includegraphics[width=0.195\linewidth]{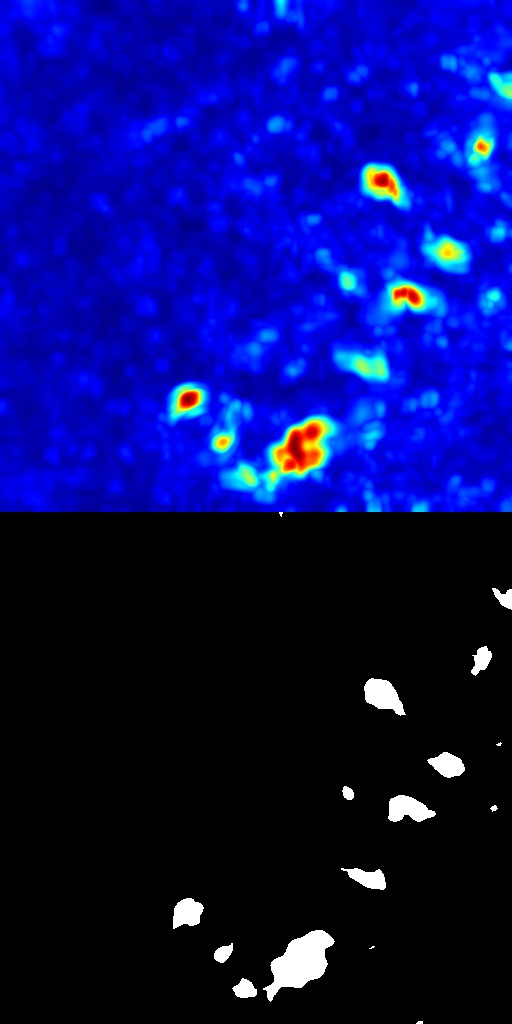}} \hfill
\subfloat{\includegraphics[width=0.195\linewidth]{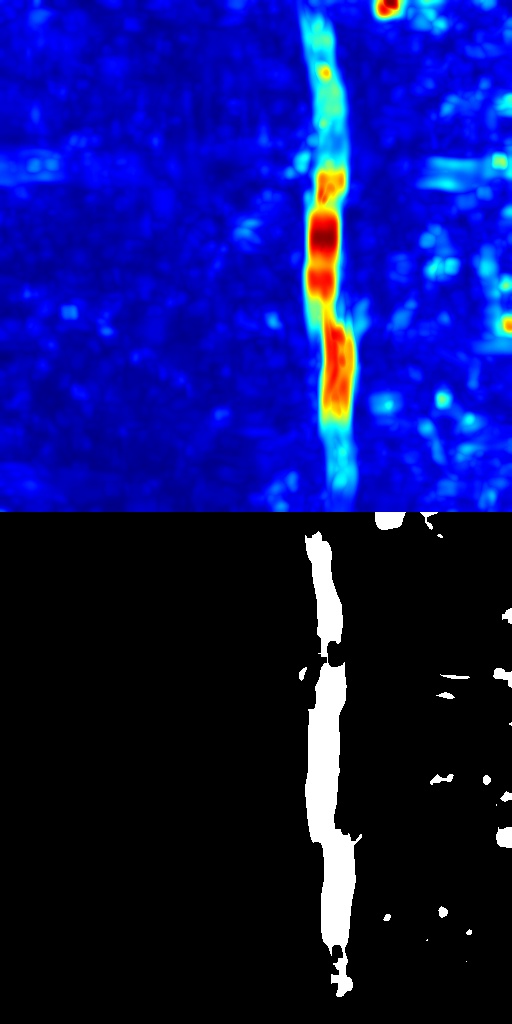}} \hfill
\subfloat{\includegraphics[width=0.195\linewidth]{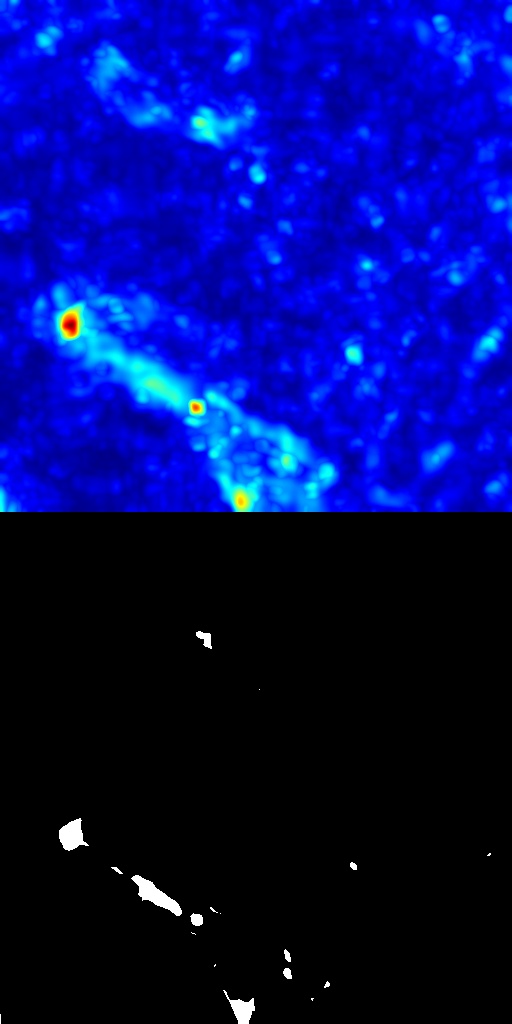}} \hfill \\[-20px]

\subfloat{\includegraphics[width=0.195\linewidth]{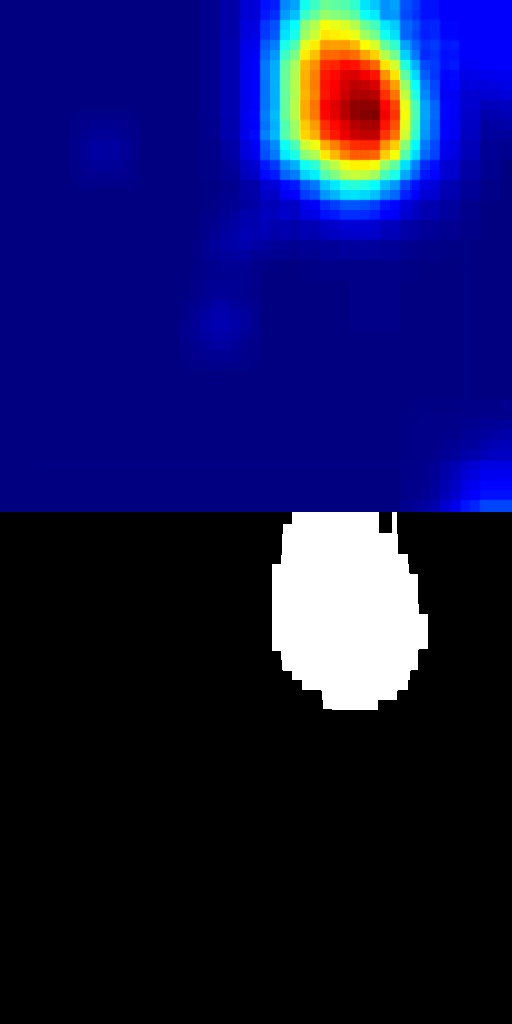}} \hfill
\subfloat{\includegraphics[width=0.195\linewidth]{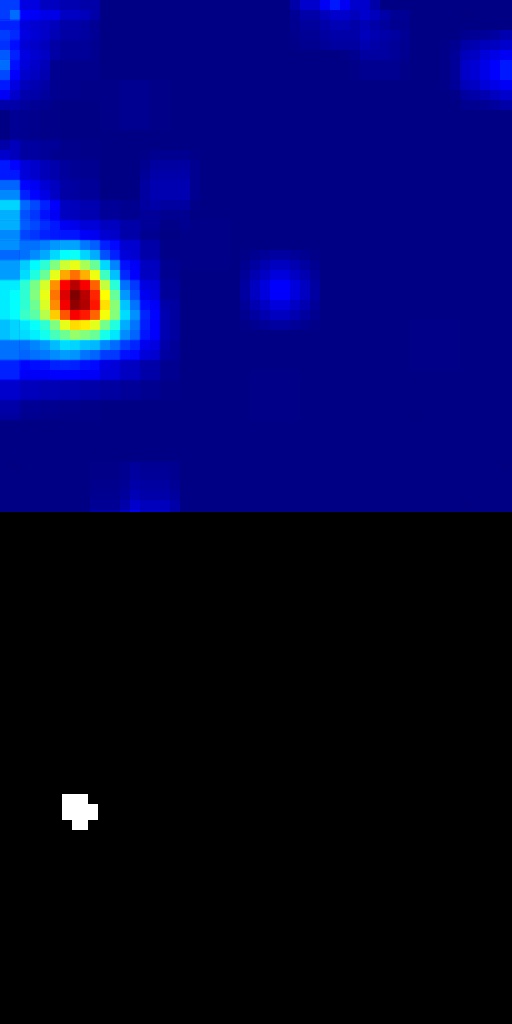}} \hfill
\subfloat{\includegraphics[width=0.195\linewidth]{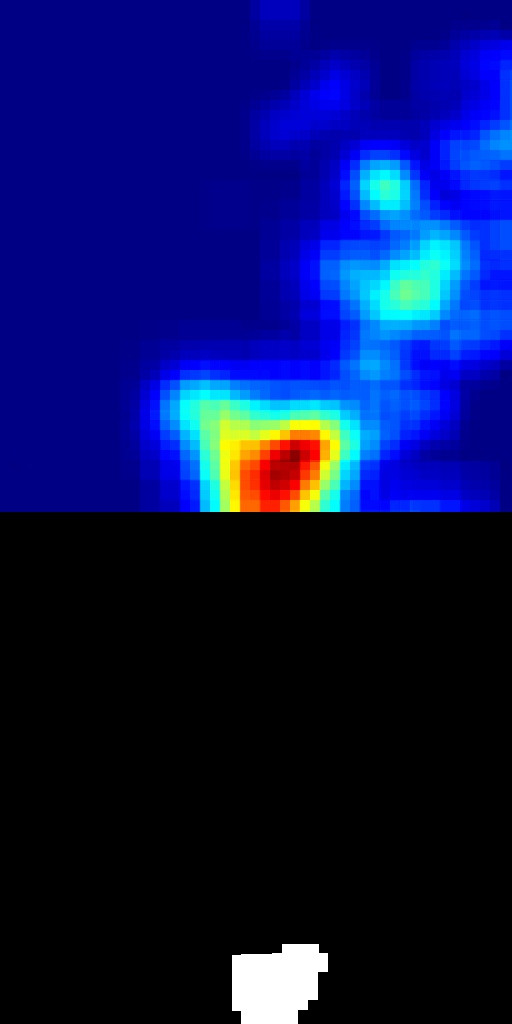}} \hfill
\subfloat{\includegraphics[width=0.195\linewidth]{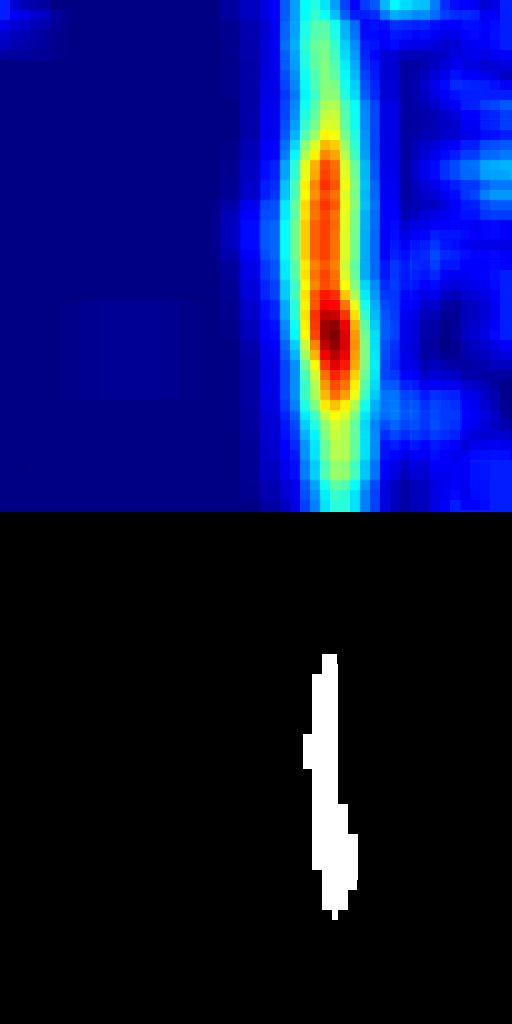}} \hfill
\subfloat{\includegraphics[width=0.195\linewidth]{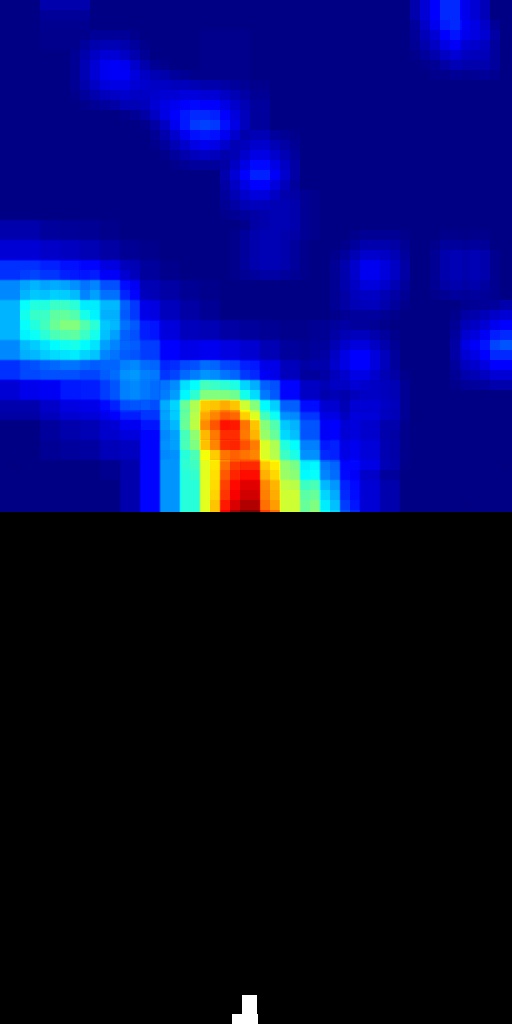}} \hfill \\[-20px]

\subfloat{\includegraphics[width=0.195\linewidth]{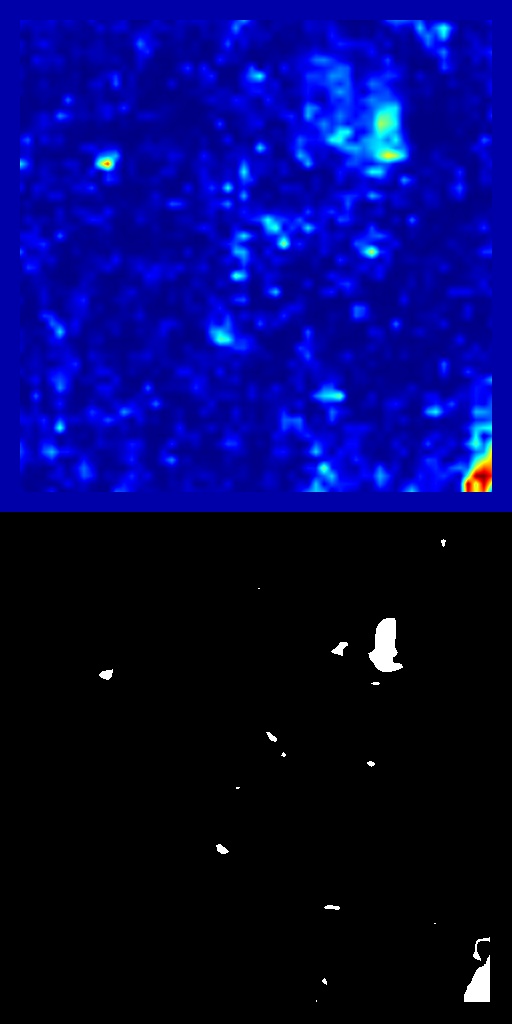}} \hfill
\subfloat{\includegraphics[width=0.195\linewidth]{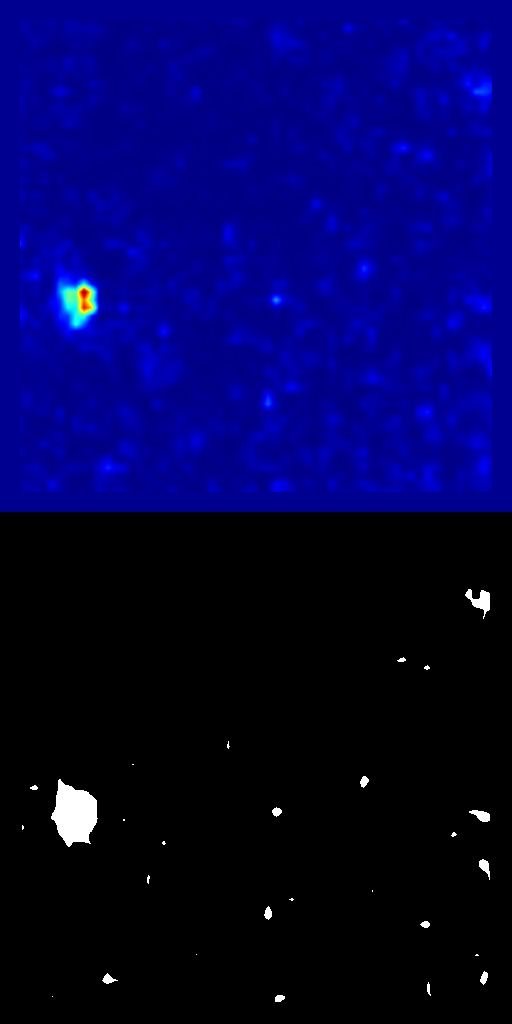}} \hfill
\subfloat{\includegraphics[width=0.195\linewidth]{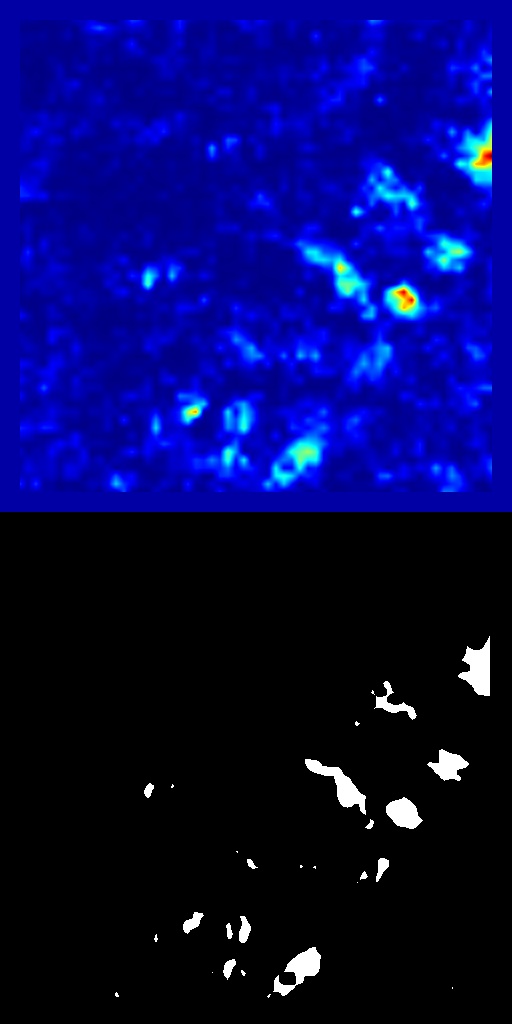}} \hfill
\subfloat{\includegraphics[width=0.195\linewidth]{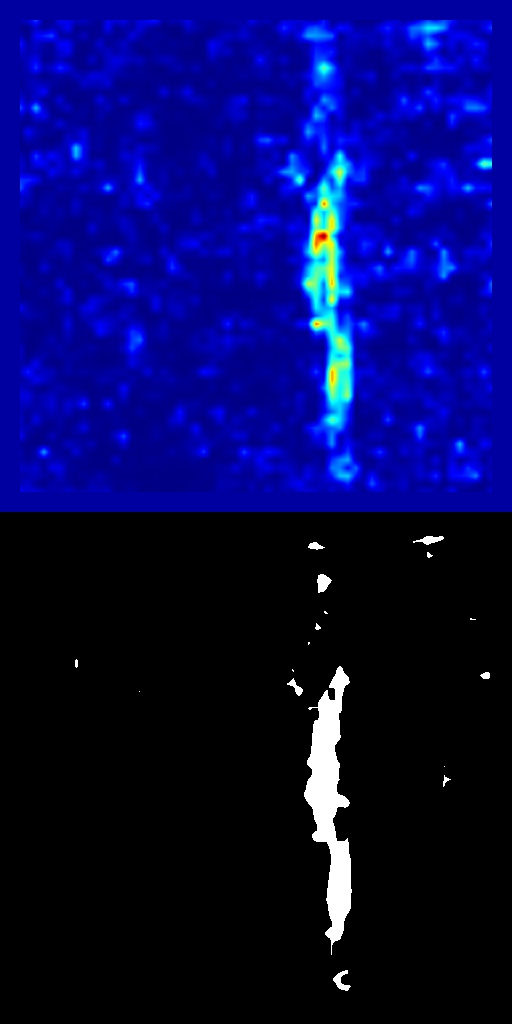}} \hfill
\subfloat{\includegraphics[width=0.195\linewidth]{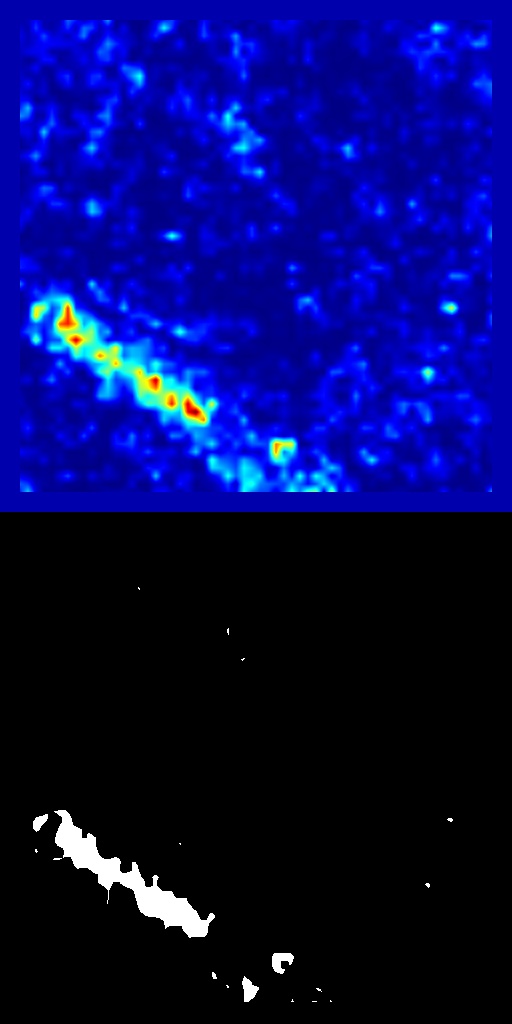}} \hfill \\[-20px]

    \caption{Results on our industrial data. First row: original diffuse image. Following rows show the anomaly score map and the segmentation with AS=0, for three variants of our proposed method: \textit{PCA+PixelNFA}, \textit{Gabor+BlockNFA}, and \textit{ResNet+PixelNFA}. All defects are detected in all cases, and AS=0 provides a good choice for the detection threshold.}
    \vspace{-10pt}
    \label{fig:results-bader}
\end{figure}